\documentclass{article}
\usepackage{spconf,amsmath,graphicx}
\usepackage{pgfplots} 
\pgfplotsset{width=8cm,compat=1.9} 
\usepgfplotslibrary{external}
\usepackage{stfloats}
\usepackage{iccv}
\usepackage{times}
\usepackage{epsfig}
\usepackage{graphicx}
\usepackage{amsmath}
\usepackage{amssymb}
\usepackage{subfig}
\usepackage{enumitem}

\title{Deep Saliency Models : The Quest For The Loss Function}
\name{Alexandre Bruckert$^{ 1}$, Hamed R. Tavakoli$^{ 2}$, Zhi Liu$^{ 3}$, Marc Christie$^{ 1}$, Olivier Le Meur$^{ 1}$}
\address{{}$^{ 1}$ Univ  Rennes, IRISA, CNRS, France\\$^{ 2}$Aalto University, Finland\\$^{ 3}$Shangai University, China}

\begin{document}
%
\maketitle
\begin{abstract}
   Recent advances in deep learning have pushed the performances of visual saliency models way further than it has ever been. Numerous models in the literature present new ways to design neural networks, to arrange gaze pattern data, or to extract as much high and low-level image features as possible in order to create the best saliency representation. However, one key part of a typical deep learning model is often neglected: the choice of the loss function. 
   
   In this work, we explore some of the most popular loss functions that are used in deep saliency models. We demonstrate that on a fixed network architecture, modifying the loss function can significantly improve (or depreciate) the results, hence emphasizing the importance of the choice of the loss function when designing a model. We also introduce new loss functions that have never been used for saliency prediction to our knowledge. And finally, we show that a linear combination of several well-chosen loss functions leads to significant improvements in performances on different datasets as well as on a different network architecture, hence demonstrating the robustness of a combined metric. 
\end{abstract}
\section{Introduction}

Despite decades of research, visual attention mechanisms of humans remain complex to understand and even more complex to model. With the availability of large databases of eye-tracking and mouse movements recorded on images~\cite{Bylinskii2015saliency,SaliconDB}, there is now a far better understanding of the perceptual mechanisms. Significant progress has been made in trying to predict visual saliency, \ie computing the topographic representation of visual stimulus strengths across an image. \emph{Deep saliency models} have strongly contributed to this progress.

However, as recently pointed out by Borji~\cite{borji2018saliency}, a neglected challenge in the design of a deep saliency model is the choice of an appropriate loss function. In~\cite{Jetley2016} a probabilistic end-to-end framework was proposed and five relevant loss functions were studied. Yet, to the best of our knowledge, none of the papers concerning the challenges in designing deep saliency models have investigated this aspect properly, despite its influence on the quality of the results. Important questions therefore arise: how do different loss functions affect performance of deep saliency networks? Which loss functions perform better than others and on which metrics? Is there actually substantial benefits in combining loss functions? And how does the combination of loss functions perform with respect to individual loss functions? In this work, we seek answers to such questions by conducting a series of extensive experiments with both well-known and newly designed loss functions. 

For this purpose, we first categorize loss functions per type of metric : (i) pixel-based comparisons \eg Mean Square Error, Absolute Errors Exponential Absolute Difference, (ii) distribution-based metrics \eg Kullback-Leibler divergence, Bhattacharya loss, binary cross-entropy, (iii) saliency-inspired metrics such as Normalized Scanpath Saliency or Pearson's Correlation Coefficient, or (iv) \emph{perceptual-based metrics}, gathering two novel metrics we propose in this paper which are inspired from image style transfer, and measure the aggregation of distances computed at each convolutional layer, between the convoluted reference image and the generated saliency map. 

We then design a novel deep saliency model to provide a fixed network architecture as a reference on which all the loss functions will be evaluated. 
Our evaluation strategy then consists in evaluating the impact of all the loss functions taken individually, on our fixed network with a fixed image dataset (MIT). Then by building on the common agreement that different  metrics favor different perceptual characteristics of the image~\cite{borji2018saliency}, we propose to further explore how the combination of loss functions, typically aggregating pixel-based, distribution-based, saliency-based and perceptual-based functions, can significantly influence the quality of the training. To demonstrate the generalization capacity of our combined metric, we measure its impact on different datasets (CAT2000 and FiWi) and also with a different network architecture (SAM-VGG).


The contributions of this paper are therefore (i) to demonstrate how the choice of the loss function can strongly improve (or depreciate) the quality of a deep saliency model, and (ii) how an aggregation of carefully selected loss functions can lead to significant improvements, both on the fixed network architecture we proposed, but also on some other architectures and datasets.

The paper is organized as follows. Section~\ref{sec:relatedWorks} presents the related works. The loss functions for training a deep architecture aiming to predict saliency map are described in Section~\ref{sec:loss}. Section~\ref{sec:experiments} presents the comprehensive analysis of loss functions and their combinations. Conclusions are drawn in the last section.

\section{Related works}\label{sec:relatedWorks}

Computational models of saliency prediction, a long standing problem in computer vision, have been studied from so many perspectives that going through all is beyond the scope of this manuscript. We, thus, provide a brief account of relevant works and summarize them in this section. We refer the readers to~\cite{Borji2013Bench,Borji_2013_study} for an overview.

To date, from a computer vision perspective, we can divide the research on computational models of saliency prediction into two era (1) pre-deep learning, and (2) deep learning. 
During the pre-deep learning period, significant number of saliency models were introduced, e.g.~\cite{Itti1998,Bruce2005,Harel2006,Hou2012,Zhang2013}, and numerous survey papers looked into these models and their properties, e.g.~\cite{Borji2013Bench,Zhao2013Bench}. During this
period the community converged into adopting eye tracking as a medium for obtaining ground truth and dealt with challenges regarding the evaluation and the models, \eg~\cite{Riche2013ICCV,Borji2013ICCV}. This era was then replaced by saliency models based on deep learning techniques~\cite{borji2018saliency}, which will be the main focus of this paper.

We therefore outline the recent research developments of deep saliency model era from two perspectives, (1) challenges of deep models and works that addressed them, and (2) the deep saliency models. We, then, stress the importance of task specific loss functions in computer vision.

\vspace{10pt}
\textbf{Challenges of deep saliency models.}

The use of deep learning introduced new challenges to the community. The characteristics of most of the models shifted towards data intensive models based on deep convolutional neural networks (CNNs). To train a model, a huge amount of data is required; motivating the search for alternatives to eye tracking databases like mouse tracking~\cite{SaliconDB}, or pooling all the existing eye tracking databases into one~\cite{Bruce2016}. 

To improve the training, Bruce~\etal~\cite{Bruce2016} investigated the factors required to take into account when relying on deep models, \eg, pre-processing steps, tricks for pooling all the eye tracking databases together and other nuances of training a deep model. Authors, however, considered only one loss function in their study.

Tavakoli~\etal~\cite{Tavakoli2017} looked into the correlation between mouse tracking and eye tracking at finer details, showing the data from the two modalities are not exactly the same. They demonstrated that, while mouse tracking is useful for training a deep model, it is less reliable for model selection and evaluation in particular when the evaluation standards are based on eye tracking.

Given the sudden boost in overall performance by saliency models using deep learning techniques, Bylinskii~\etal~\cite{Bylinskii2016} reevaluated the existing benchmarks and  looked into the factors influencing the performance of models in a finer detail. They quantified the remaining gap between models and human. They argued that pushing performance further will require high-level image understanding.

Recently Sen~\etal~\cite{He2019UnderstandingAV} investigated the effect of model training on neuron representations inside a deep saliency model. They demonstrated that (1) some visual regions are more salient than others, and (2) the change in inner-representations is due to the task that original model is trained on prior to being fine-tuned for saliency.

\vspace{10pt}
\textbf{Deep saliency models.}

The deep saliency models fall into two categories, (1) those using CNNs as a fixed feature extractors and learn a regression from feature space into saliency space using a none-neural technique, and (2) those that train a deep saliency model end-to-end. The number of models belonging to the first category is limited. They are not comparable within the context of this research because the regression is often carried out such that the error can not be back-propagated, \eg, \cite{Vig2014} employs support vector machines and \cite{R.Tavakoli2017b} uses extreme learning machines. Our focus is, however, the second group.

Within end-to-end deep learning techniques, the main research has been on architecture design. Many of the models borrow the pre-trained weights of an image recognition network and experiment combining different layers in various ways. In other words, they engineer an encoder-decoder network that combines a selected set of features from different layers of a recognition network. In the following we discuss some of the most well-known models.
 
Huang~\etal~\cite{Huang2015ICCV} proposed a multi-scale encoder based on VGG networks and learns a linear combination from responses of two scales (fine and coarse). K\"{u}mmerer~\etal~\cite{Kummerer2014b} use a single scale model using features from multiple layers of AlexNet.
Similarly, K\"{u}mmerer~\etal~\cite{Kummerer2017b} and Cornia~\etal~\cite{Cornia2016} employed single scale models with features from multiple layers of a VGG architecture.

There has been also a wave of models incorporating recurrent neural architectures. Han and Liu~\cite{liu2018deep} proposed a multi-scale architecture using convolutional long-short-term memory (ConvLSTM). It is followed by \cite{cornia2018predicting} using a slight modified architecture using multiple layers in the encoder and a different loss function. Recurrent models of saliency prediction are more complex than feed-forward models and
more difficult to train. Moreover, their performance is not yet significantly better than some recent feed-forward networks such as EML-NET~\cite{Jia2018}.

In the literature of deep saliency models, a loss function or a combination of several ones is chosen based on intuition, expertise of the authors or sometimes mathematical formulation of a model. K\"{u}mmerer \etal~\cite{Kuemmerer2015a} introduces the idea that information-theory can be a good inspiration for saliency metrics. They use the information gain to explain how well a model performs compared to a gold-standard baseline. Consequently, they use the log-likelihood for a loss function in~\cite{kummerer2016deepgaze}, achieving state-of-the-art results in saliency prediction.  Jetley \etal~\cite{Jetley2016} are part of the very few who specifically focused on the design of a loss functions for saliency models. They proposed the use of Bhattacharyya distance and compared it to 4 other probability distances. In this paper, in contrast to~\cite{Jetley2016}, we (1) adopt a principled approach to compare existing loss functions and their combinations and (2) investigate their convergence properties over different datasets and network architectures. 

\vspace{10pt}
\textbf{Deep Learning, loss functions and computer vision.}

With the application of deep learning techniques to computer vision domain, 
the choice of appropriate loss function for a task has become a critical aspect of the model training. The computer vision community have been successful in developing task tailored loss functions to improve a model, \eg, encoding various geometric properties for pose estimation~\cite{kendall2017posenet}, curating loss functions enforcing perceptual properties of vision for various generative deep models~\cite{Johnson2016Perceptual}, exploiting the sparsity within the structure of problem, \eg, class imbalanced between background and foreground in detection problem, for reshaping standard loss functions and form a new effective loss functions~\cite{focalloss}. Our efforts follows the same path to identify the effectiveness of a range of loss functions in saliency prediction. 


\section{Loss functions for deep saliency network}\label{sec:loss}
Before delving into the description of loss functions, we present the architecture of the convolutional neural network that will be used throughout this paper. After this presentation, we elaborate on the tested loss functions.

\subsection{Proposed baseline architecture}\label{subsection:network}
\begin{figure}[htbp]
\begin{center}
   \includegraphics[width=\linewidth]{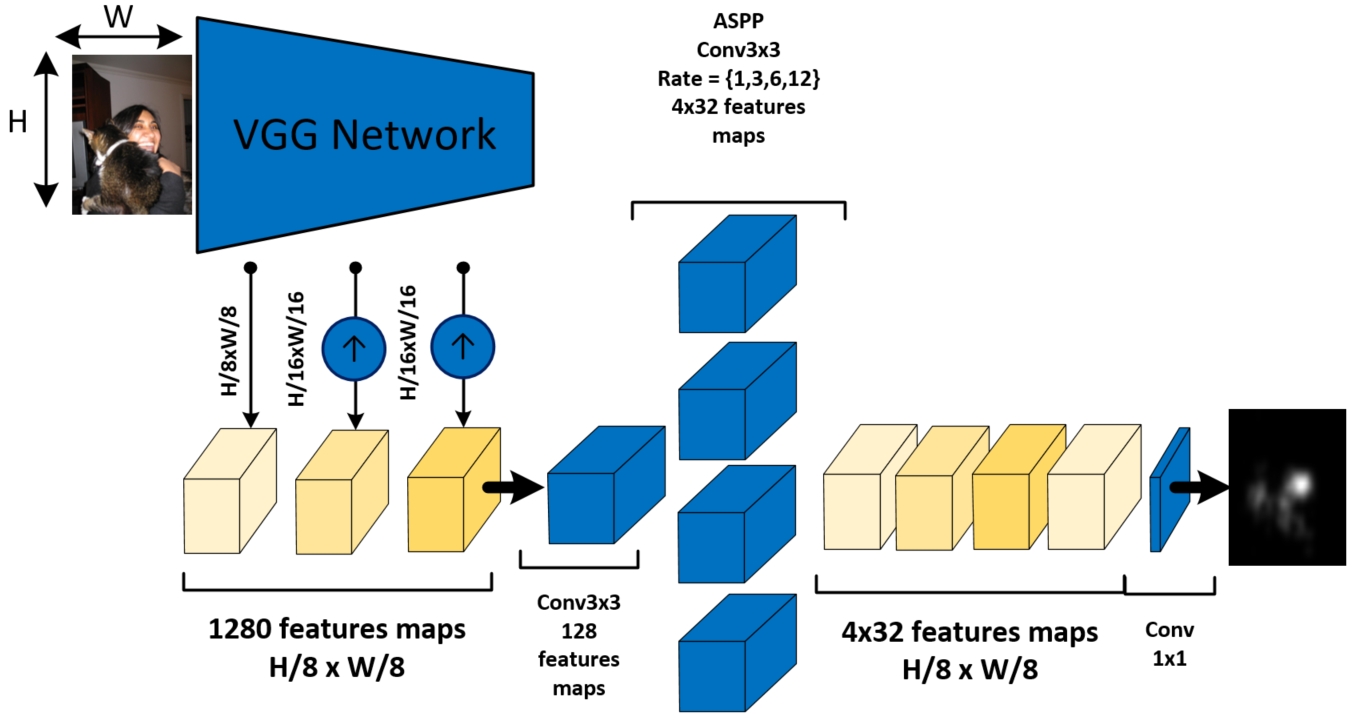}
\end{center}
   \caption{Architecture of the proposed deep network.}
\label{fig:architecture}
\end{figure}

Figure~\ref{fig:architecture} presents the overall architecture of the proposed model. The purpose of designing a new architecture is only to perform a comparison with existing architectures. Our architecture is based on the deep gaze network of~\cite{kummerer2016deepgaze} and on the multi-level deep network of~\cite{Cornia2016}. The pre-trained VGG-16 network~\cite{Simonyan2014very} is used for extracting deep features of an input image ($400\times 300$) from layers conv$3$\_pool, conv$4$\_pool, conv$5$\_conv3. Feature maps of layers conv$4$\_pool and conv$5$\_conv3 are rescaled to get feature maps with a similar spatial resolution.\\
That feature map with 1280 channels is then fed into a shallow network composed of the following layers: a first  convolutional layer allows us to reduce by a factor ten the number of channels, which are then processed by an ASPP (an atrous spatial pyramid pooling~\cite{Chen2017rethinking}) of 4 levels. Each level has a convolution kernel of $3\times 3$, a stride equal to 1 and a depth of 32. The dilatation rates are 1, 3, 6, and 12. The ASPP benefit is to catch information in a coarse-to-fine approach while keeping the resolution of the input feature maps. The output of the four pyramid levels are then merged together, \ie leading to $4\times 32$ maps. The last $1\times 1$ convolutional layer reduces the data dimensionality to 1 feature map. This map is then smoothed by a Gaussian filter $5\times 5$ with a standard deviation of 1. The activation function of these layers is a ReLU activation.\\
The network was trained over the MIT dataset composed of more than 1000 images~\cite{Judd2009learning}. We split this dataset into 500 images for the training, 200 images for the validation and the rest for the test. We use a batch size of 60, and the stochastic gradient descent. To prevent over-fitting, a dropout layer, with a rate of 0.25, is added on top the network. The learning rate is set to 0.001. During the training, the network was validated against the validation test to monitor the convergence and to prevent over-fitting. The number of trainable parameters is approximately 1,62 millions. In the following section, we present the different tested loss functions used during the training phase.

\subsection{Loss functions}
Let $\mathcal{I}:\Omega\subset\mathcal{R}^2\mapsto \mathcal{R}^3$ an input image of resolution $N\times M$. We suppose $S$ and $\hat{S}$ the vectorized human and the predicted saliency maps, \ie $S$ and $\hat{S}$ are in $\mathcal{R}^{N\times M}$. Let also $S^{fix}$ be the human eye fixations map, \ie a $N\times M$ image
with $1$ or $0$ pixels. In the following, we present thirteen loss functions $\mathcal{L}$ tested in this study. There are classified into four categories according to their characteristics: pixel-based, distribution-based, saliency-inspired and perceptual-based.

\subsubsection{Pixel-based loss functions}
For pixel-based loss functions $\mathcal{L}$, we assume that $S$ and $\hat{S}$ are in $\left[0,1\right]$. We evaluate the following loss functions:
\begin{itemize}[leftmargin=*]
    \item Mean Squared Error (MSE) measures the averaged squared error between prediction and ground truth: 
    \begin{equation}
        \mathcal{L}(S,\hat{S})=\frac{1}{N\times M}\sum_{i=1}^{N\times M} (S_i-\hat{S}_i)^2
    \end{equation}
    \item Exponential Absolute difference (EAD):
    \begin{equation}
        \mathcal{L}(S,\hat{S})=\frac{1}{N\times M}\sum_{i=1}^{N\times M} \left(exp(|S_i-\hat{S}_i|)-1\right)
    \end{equation}
    \item Absolute Error (AE):
    \begin{equation}
        \mathcal{L}(S,\hat{S})=\frac{1}{N\times M}\sum_{i=1}^{N\times M} |S_i-\hat{S}_i|
    \end{equation}
    \item Weighted MSE Loss (W-MSE):
    \begin{equation}
        \mathcal{L}(S,\hat{S})=\frac{1}{N\times M}\sum_{i=1}^{N\times M}  w_i\left(S_i-\hat{S}_i\right)^2
    \end{equation}
    The weight $w_i$ allows to put more emphasis on errors occurring on salient pixels of the ground truth $S$. Two functions are tested in this paper. 
    In~\cite{Cornia2016}, authors defined the loss function (MLNET): $w_i=\frac{1}{\alpha-S_i}$ and $\alpha=1.1$. Therefore, when $S_i=1$, the error is multiplied by a factor 10, whereas when $S_i=0$, the multiplying factor is equal to $0.90$.  We also consider a weighting function based on a parametric sigmoid function (SIG-MSE): $w_i=\frac{k}{1+\exp(-k\times(S_i-\lambda))}$, where $k=10$ and $\lambda$ varies between 0 and 1.
\end{itemize}


\subsubsection{Distribution-based loss functions}
For the distribution-based loss functions, we consider that the vectorized human and the predicted saliency pixels represent a probability to be salient. For that the network presented in section~\ref{subsection:network} is modified in order to output pixel-wise predictions that can be considered as probabilities for independent binary random variables~\cite{Pan2017salgan}. An element-wise sigmoid activation function is then added as being the last layer. The following loss functions are investigated:

\begin{itemize}[leftmargin=*]
    \item Kullback-Leibler divergence (KLD) measures the divergence between the distribution $S$ and $\hat{S}$:
    \begin{equation}
        \mathcal{L}(S,\hat{S})=\sum_{i=1}^{N\times M} \hat{S}_i\log\frac{\hat{S}_i}{S_i}
    \end{equation}
    \item Bhattacharya loss (BHAT)  measures the similarity between the distribution $S$ and $\hat{S}$:
    \begin{equation}
        \mathcal{L}(S,\hat{S})=\sum_{i=1}^{N\times M} \sqrt{S_i\hat{S}_i}
    \end{equation}
    \item Binary Cross Entropy (BCE) assumes that the saliency prediction $\hat{S}$ as well as the ground truth saliency map $S$ are composed of independent binary random variables:
    \begin{equation}
        \mathcal{L}(S,\hat{S})=-\sum_{i=1}^{N\times M} \left(S_i\log\hat{S}_i + (1-S_i)\log(1-\hat{S}_i)\right)
    \end{equation}
    \item Weighted Binary Cross Entropy (W-BCE): compared to the BCE loss, a global weight $w$ is introduced to consider that there are much more non salient areas than salient areas~\cite{Wang2018video}. It allows to put more emphasis on errors occurring when $S\rightarrow 1$ and $\Hat{S}\rightarrow 0$ ($w>>0.5$) or when $S\rightarrow 0$ and $\hat{S}\rightarrow 1$ ($w<<0.5$): 
    {
    \begin{eqnarray}\nonumber
        \mathcal{L}(S,\hat{S}) &=& -\sum_{i=1}^{N \times M} \left( w\times S_i\log\hat{S}_i \right.\\ 
        && \left. + (1-w)(1-S_i)\log(1-\hat{S}_i) \right)
    \end{eqnarray}
    }
    \normalsize
    \item Focal Loss (FL) : In order to deal with the large foreground-background class imbalance encountered during the training of dense detectors, Lin\etal~\cite{focalloss} modified the binary cross entropy loss function. Such class imbalance is also relevant in the context of saliency prediction, for which the  ground truth saliency map mainly consists of null or close to zero, creating a similar phenomenon. The approach is quite similar to W-BCE, except that the weight is locally adjusted and based on a tunable $\gamma$ power of the predicted saliency. As in~\cite{focalloss}, we set by default the $\gamma$ value equal to 2:
    \begin{eqnarray}\nonumber
        \mathcal{L}(S,\hat{S})&=&-\sum_{i=1}^{N \times M} \left( (1-\hat{S}_i^{\gamma})\times S_i\log\hat{S}_i \right. \\
        && \left. + \hat{S_i}^{\gamma}(1-S_i)\log(1-\hat{S}_i)  \right)
    \end{eqnarray}
    \normalsize
    \item Negative Logarithmic Likelihood (NLL): As shown by K\"{u}mmerer \etal~\cite{Kuemmerer2015a}, information theory provides strong insights when it comes to saliency models. For instance, in \cite{kummerer2016deepgaze}, they use maximum likelihood learning. Let $I$ be the set of fixations in an image, and $N_i$ the number of those fixations. The logarithm of the prediction at the coordinates of each fixation is then computed :
        \begin{equation}
        \mathcal{L}(\hat{S})=-\frac{1}{N_i}\sum_{i\in I} \left(\log\hat{S}_i\right)
    \end{equation}
\end{itemize}

\subsubsection{Saliency-inspired loss functions}
Saliency predictions are usually evaluated using several metrics~\cite{LeMeur2013}. Those metrics are good candidates to use as loss functions, since they capture several properties that are specific to saliency maps.

\begin{itemize}[leftmargin=*]
    \item Normalized Scanpath Saliency (NSS): This metric was introduced in~\cite{peters2005nss}, to evaluate the degree of congruency between human eye fixations and a predicted saliency map. Instead of relying on a saliency map as ground truth, the predictions are evaluated against the true fixations map.  The value of the saliency map at each fixation point is normalized with the whole saliency map variance:
    \begin{equation}
        \mathcal{L}(S^{fix},\hat{S})=\frac{1}{N \times M} \sum_{i = 1}^{N \times M} \frac{\hat{S}_i - \mu(\hat{S}_i)}{\sigma(\hat{S}_i)}S^{fix}_i
    \end{equation}
    \item  Pearson's Correlation Coefficient (CC) measures the linear correlation between the ground truth saliency map and the predicted saliency map :
    \begin{equation}
        \mathcal{L}(S,\hat{S})= \frac{\sigma(S, \hat{S})}{\sigma(S)\sigma(\hat{S})}
    \end{equation}
    where $\sigma(S, \hat{S})$ is the covariance of $S$ and $\hat{S}$.
\end{itemize}    

\subsubsection{Perceptual-based loss functions}
We propose two new loss functions for deep saliency, that have been applied with success in image style-transfer problems~\cite{Johnson2016Perceptual, gatys2015gram}. The idea is to compare the representations of the ground truth and predicted saliency maps that are extracted from different layers of a fixed pre-trained convolutional neural network. The idea behind those losses is to take into account not only the saliency map, but also the deep hidden patterns that could exist, as well as the potential relationship between such patterns. Let $\phi_j(S)$ be the activation at the $j^{th}$ layer of the VGG network when fed a saliency map $S$. $\phi_j(S)$ is then of size $C_j \times H_j \times W_j$, where $C_j$ represents the number of filters, $H_j $ and $W_j$ represent the height and width of the feature maps at the layer $j$, respectively. We also denote $J$ the set of layers from which we extract the representations. In this work, we extracted the outputs of the 5 pooling layers of a fixed VGG-16 network~\cite{Simonyan2014very} pre-trained on the ImageNet dataset, representing a total of 1920 filters:

\begin{itemize}[leftmargin=*]
    \item Deep Features loss (DF) measures the Euclidean distance between the feature representations:
    \begin{equation}
        \mathcal{L}(S,\hat{S})=\sum_{j \in J} \frac{1}{C_j \times H_j \times W_j} \|\phi_j(S) - \phi_j(\hat{S}) \|^2
    \end{equation}
    \item Gram Matrices of Deep Features loss (GM) : In order to leverage the potential statistical dependency between features maps, we propose a new loss relying on Gram matrices. For this purpose, we reshape the output $\phi_j(S)$ into a matrix $\psi$ of size $C_j \times (H_jW_j)$. Then, for each layer $j$, the Gram matrix $G_{j}^{\phi}(S)$ of size $C_j \times C_j$ is defined as follows:
    \begin{equation}
        G_{j}^{\phi}(S) = \frac{1}{C_j \times H_j \times W_j} \psi \psi^\top
    \end{equation}
    The loss function is then the sum of the squared Frobenius norm of the difference between the Gram matrices $G_{j}^{\phi}$, $j\in J$:
    \begin{equation}
        \mathcal{L}(S, \hat{S})=\sum_{j \in J} \| G_{j}^{\phi}(S) - G_{j}^{\phi}(\hat{S}) \|^2_F
    \end{equation}
\end{itemize}


\subsubsection{Center-bias regularization}
\begin{figure}[htbp]
    \centering
    \includegraphics[width=\linewidth]{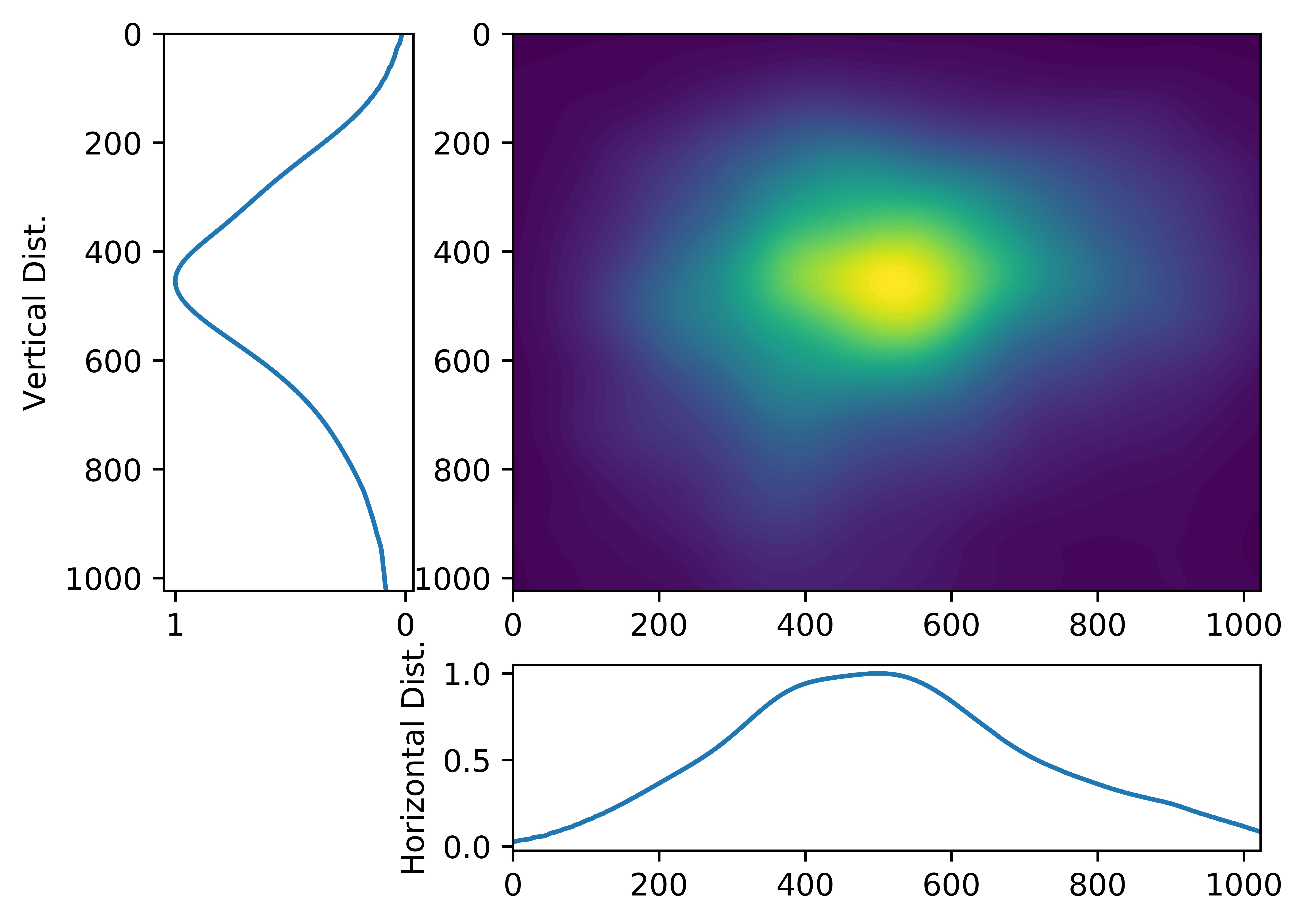}
    \caption{Averaged colored saliency map of the training part of MIT dataset. Horizontal and vertical marginal distributions are also plotted, illustrating the center bias.}
    \label{fig:center_bias}
\end{figure}

Since our model does not take into account the center-bias with a learned prior, like in \cite{cornia2018predicting, Cornia2016}, we add a regularization term. We compute a center-bias map $B$ as the mean of the ground truth maps from the training part of the MIT dataset, and add  to the loss function the regularization term $R_i$ for each pixel $i$:
\begin{equation}
    R_i=\alpha ( \hat{S}_i - B_i )^2
\end{equation}
We empirically set the parameter $\alpha$ to $0.1$, even though it could be optimized to improve final results. This regularization will later be referred as R in Table~\ref{table:performance}. The center-bias map $B$ is illustrated in Figure~\ref{fig:center_bias}. 

\subsubsection{Linear combinations}
All presented loss functions evaluate  different characteristics of the predicted saliency maps. We can then hypothesize that a linear combination of some of those loss functions could lead to better results, as it would aggregate the particularities of all measures (a strategy already adopted by \cite{cornia2018predicting}). We decided to evaluate two linear combinations: the first one (LC 1) combining KLD, CC and NSS loss functions, and the second one (LC 2) adding Deep Features loss, Gram Matrices loss and sigmoid-weighted MSE. This specific combination was chosen because it relies on an existing successful combination and also aggregates the four types of metrics together. 

We followed the work of~\cite{cornia2018predicting} to set the coefficients for the first linear combination: $-1$ for the NSS, $-2$ for the CC, and $10$ for the KLD. In the LC 2 combination, we kept those coefficients and set arbitrarily all the other ones as 1. We used the best $\lambda$ parameter that we tested for the sigmoid-weighted MSE in the linear combination ($\lambda = 0.55$).

\section{Experiments}\label{sec:experiments}
\subsection{Testing protocols}
To carry out the evaluation, we use seven quality metrics applied on the MIT benchmark~\cite{Bylinskii2015saliency,LeMeur2013}: CC (correlation coefficient, $CC\in\left[-1,1\right]$), SIM (similarity, intersection between histograms of saliency, $SIM\in\left[0,1\right]$), AUC (Area Under Curve, $AUC\in\left[0,1\right]$), NSS (Normalized Scanpath Saliency, $NSS\in\left]-\infty,+\infty\right[$), EMD (Earth Mover Distance, $EMD\in\left[0,+\infty\right[$) and KL (Kullback Leibler divergence, $KL\in\left[0,+\infty\right[$).\\
The similarity degree between prediction and ground truth is computed over 299 images. 

\subsection{Performance of the proposed model}
Table~\ref{table:performance_proposed_model_MSE} presents the performance of the proposed model (when trained with the MSE loss function) compared to existing models, \ie Itti~\cite{Itti1998}, Rare2012~\cite{Riche2013ICCV}, GBVS~\cite{Harel2006}, AWS~\cite{GarciaDiaz2012}, Sam-ResNet \& Sam-VGG~\cite{cornia2018predicting}, SalGan~\cite{Pan2017salgan}, ShallowNet \& DeepConvNet~\cite{pan2016shallow}. All models are evaluated on the test dataset as defined in Section~\ref{subsection:network}.

\begin{table}[ht!]
\small
\caption{Performance of our model. Best performances are in bold. (AUC-B=AUC-Borji; AUC-J=AUC-Judd)}
\label{table:performance_proposed_model_MSE}
\centering
\begin{tabular}{lccccc}
\hline
            & CC~$\uparrow$	& SIM~$\uparrow$	& AUC-J~$\uparrow$	& AUC-B~$\uparrow$	 & NSS~$\uparrow$	\\
\hline \hline
Itti	        &0.28	&0.35	&0.71	&0.71	&0.79\\
Rare2012	    &0.46	&0.43	&0.78	&0.79	&1.40\\
GBVS	        &0.49	&0.44	&0.81	&0.81	&1.38\\
AWS	            &0.39	&0.40	&0.75	&0.76	&1.22\\
\hline
Sam-ResNet	    &0.68	&\textbf{0.60}	&0.85	&0.79	&\textbf{2.43}\\
SalGan	        &\textbf{0.70}	&0.58	&\textbf{0.86}	&\textbf{0.86}	&2.17\\
Sam-VGG	        &0.64	&0.57	&0.85	&0.78	&2.19\\
ShallowNet	    &0.59	&0.47	&0.83	&0.84	&1.62\\
DeepConvNet	    &0.60	&0.49	&0.84	&0.85	&1.73\\
\hline
\textbf{Our model}	    &0.64	&0.45	&0.81	&0.84	&2.04\\
\textit{(ranking)} &  \textit{3/10} & \textit{5/10} & \textit{6/10} & \textit{3/10} & \textit{4/10} \\
\hline
\end{tabular}
\end{table}
According to the evaluation, the proposed model performs rather well and is in the top 4 models. The best performing models are Sam-ResNet and SalGan. Do note however that other models are not specifically trained on the MIT dataset, unlike ours.

\subsection{Loss function performance}
Table~\ref{table:performance} presents the performance obtained with the different loss functions. 

\vspace{10pt}
\textbf{Which category of loss functions provide the best performances?} 

Except the perceptual-based loss functions, results suggest that the pixel-based, the distribution-based and the saliency-inspired loss functions perform similarly. However, the perceptual-based loss functions  we introduced, namely DF and GM, do not perform well individually compared to the aforementioned losses. It might be due to the feature maps used for these losses which are extracted using a deep convolutional network that was trained using natural images, and not saliency maps. The representation of the saliency maps in the feature space might then not be appropriate for this task.

\vspace{10pt}
\textbf{Designing a stronger loss from weaker losses.} 

Results also suggest that a simple linear combination of well known loss functions increases the ability of the network to predict saliency maps. While keeping the number of trainable parameters unchanged, we succeed in improving up to 14\% the correlation coefficient when we compared the best linear combination  ($CC=0.7291$) to the classical MSE ($CC=0.6388$). In a more general way, linear combinations of the loss functions systematically improve the results on most of the metrics. Such fluctuations between the performances of the different loss functions confirm our hypothesis that the choice of the loss function is a critical part of designing a deep saliency model. Moreover, the aggregated loss of KL + CC + NSS + DF + GM improves the SIM, AUC-B and KL scores compared to the aggregation KL + CC + NSS. This reveals the influence of perceptual-based losses (DF+GM) in deep saliency models, and probably calls for future work in this direction.

\begin{figure*}[htbp]
    \centering
    \includegraphics[width=\linewidth]{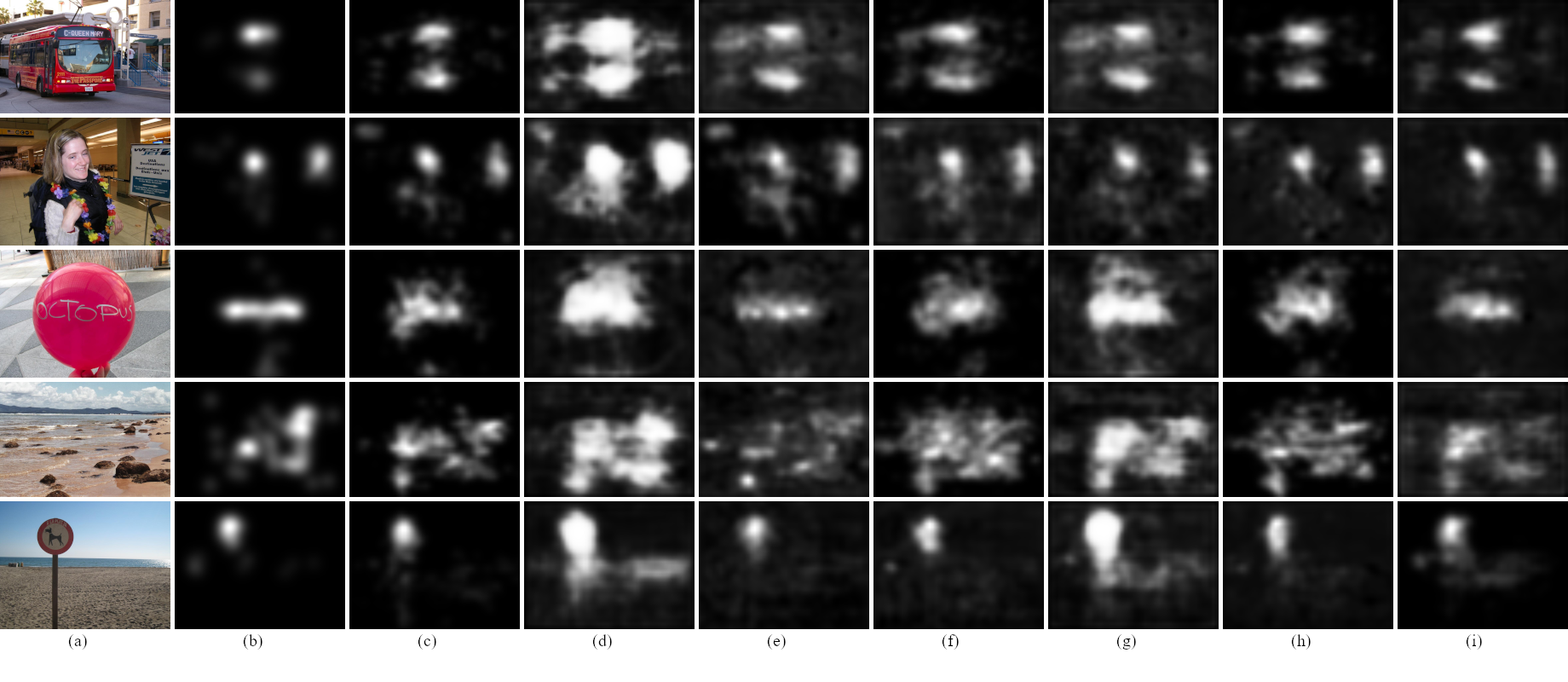}
    \caption{(a) Visual stimulus; (b) Ground truth saliency map; (c) KLD + CC + NSS + DF + GM + SIG-MSE + R combination; (d) NSS; (e) KL + CC + NSS + R; (f) MLNET-MSE; (g) KLD; (h) SIG-MSE ($\lambda = 0.55)$; (i) CC.}
    \label{fig:saliency_examples}
\end{figure*}

Beyond this quantitative analysis, Figure~\ref{fig:saliency_examples} illustrates predicted saliency maps obtained for some of the tested loss functions. Qualitatively speaking, the saliency maps obtained when using the combined loss KLD + CC + NSS + DF + GM + SIG-MSE + R look very similar to the ground truth maps. For instance, they are very condensed around the salient regions, with little noise.

\begin{table*}[ht!]
\caption{Performance of the loss functions. Best performances are in bold and the second and third best performances are in italic. R represents the center-biais regularization. (AUC-B=AUC-Borji; AUC-J=AUC-Judd)}
\label{table:performance}
\centering
\begin{tabular}{lccccccc}
\hline
            & CC~$\uparrow$	& SIM~$\uparrow$	& AUC-J~$\uparrow$	& AUC-B~$\uparrow$	 & NSS~$\uparrow$	& EMD~$\downarrow$	& KL~$\downarrow$\\
\hline \hline
Pixel-based loss functions\\
\hline
MSE	    &0.6388	&0.4492	&0.8118	&0.8363	&2.0580	&2.3279	&0.9472\\
EAD 	&0.6725	&0.4790	&0.8326	&0.8428	&2.2133	&2.1744	&0.8592\\
AE	    &0.6426	&0.4443	&0.8145	&0.8322	&2.1388	&2.3782	&0.9616\\
MLNET-MSE	&0.6904	&\textbf{0.5962}	&0.8416	&0.8245	&\textit{2.2468}	&\textbf{1.3581}	&1.6042\\
SIG-MSE ($\lambda=0.25$)	&0.6929	&\textit{0.5896}	&\textit{0.8512}	&0.8438	&2.1897	&\textit{1.4120}	&0.9876\\
SIG-MSE ($\lambda=0.55$)	&0.6725	&0.5646	&0.8505	&0.8542	&2.0799	&\textit{1.5137}	&0.8478\\
SIG-MSE ($\lambda=0.75$)	&0.6440	&0.5280	&\textit{0.8512}	&0.8637	&1.9480	&1.7100	&0.7686\\
\hline
Distribution-based loss functions\\
\hline
BCE	            &0.6616	&0.4712	&0.8231	&0.8380	&2.1229	&2.2600	&0.8899\\
W-BCE $w=0.9$	&0.6333	&0.4287	&0.8308	&0.8519	&1.8734	&2.4793	&1.0003\\
W-BCE $w=0.8$	&0.6363	&0.4273	&0.8305	&0.8531	&1.8909	&2.5026	&1.0067\\
W-BCE $w=0.7$	&0.6409	&0.4308	&0.8179	&0.8396	&1.9636	&2.4824	&0.9976\\
W-BCE $w=0.6$	&0.6478	&0.4335	&0.8182	&0.8412	&2.0135	&2.4800	&0.9862\\
W-BCE $w=0.5$	&0.6739	&0.4305	&0.8420	&\textbf{0.8582}	&2.0911	&2.4713	&0.9871\\
W-BCE $w=0.4$	&0.6443	&0.3992	&0.8301	&0.8468	&2.0166	&2.6625	&1.0949\\
Focal Loss      &0.6530 &0.4294 &0.8197 &0.8403 &1.9552 &2.4839 &0.9738\\
KLD	&0.6326	&0.4893	&0.8356	&0.8541	&1.7913	&2.0609	&0.8336\\
Bhat	&0.6203	&0.5029	&0.8429	&0.8567	&1.7321	&1.9209	&\textbf{0.7909}\\
NLL &0.6251 &0.4973 &0.8407 &\textit{0.8559} &1.7856 &1.8734 &\textit{0.7955}\\
\hline
Saliency-inspired loss functions\\
\hline
CC              &0.6943 &0.4994 &0.8411 &0.8386 & 1.8201 &2.1378
&0.9157\\
NSS             &0.6740 &0.4325 &0.8397 &0.8216 &\textbf{2.3142} &2.9964
&1.3498\\
\hline
Perceptual-based loss functions\\
\hline
Deep Features (DF)      &0.6065 &0.4772 &0.8308 &0.8259 &1.9731 &3.7546
&0.9675\\
Gram Matrices (GM)      &0.5911 &0.4964 & 0.8371 &0.8312 &1.8357 &2.0455 &1.1993\\
\hline
Linear combinations\\
\hline
SIG-MSE + R       &0.6813 &0.5611 &0.8507 &0.8373 &1.9734 &3.1471 &0.8349\\
KLD + CC + NSS       &\textit{0.7288} &0.5754 &\textit{0.8512} &0.8487 &\textit{2.2464} &2.1340 &0.9571\\
KLD + CC + NSS + DF + GM    &\textit{0.7192} &0.5790 &0.8492 &0.8536 &1.9652 &2.3101 &0.9387\\
KLD + CC + NSS + R       &0.7176 &0.5683 &\textit{0.8579} &0.8520 &2.2147 &2.8808 &0.8912\\
KLD + CC + NSS + DF + GM + SIG-MSE + R       &\textbf{0.7291} &\textit{0.5817} &\textbf{0.8585} &\textit{0.8563} &2.2094 &2.5517 &\textit{0.8010}\\
\hline
\end{tabular}
\end{table*}

\subsection{Does the best loss generalize well over different datasets and a different architecture?}
In this section, we test how well the best loss generalizes over two datasets, \ie CAT2000 and FiWi, and one architecture, \ie SAM-VGG. 

\vspace{10pt}
\textbf{CAT2000 and FiWi datasets.} 

CAT2000 eye tracking dataset~\cite{borji2015cat2000} is composed of 2000 images belonging to 20 different categories whereas FiWi dataset~\cite{Shen2014} is composed of more than 140 screen shots of webpages. 
Performances are given in Table~\ref{table:performance_proposed_model_CAT2000_FIWI}. Results indicate that the loss function based on the linear combination of KLD, CC, NSS, DF, GM, SGI-MSE and R allows to significantly increase the ability to predict visual saliency. Compared to the MLNET-MSE loss function, the gain in terms of CC is 16.3$\%$ and 7.1$\%$ for CAT2000 and FiWi datasets, respectively.

\begin{table}[ht!]
\small
\caption{Performance of proposed model over CAT2000 and FiWi datasets with MLNET-MSE (W-MSE), KLD + CC + NSS + R (LC 1) and KLD + CC + NSS +DF +GM + SIG-MSE ($\lambda = 0.55)$ + R (LC 2). Best performances are in bold. (AUC-B=AUC-Borji; AUC-J=AUC-Judd)}
\label{table:performance_proposed_model_CAT2000_FIWI}
\centering
\begin{tabular}{lccccc}
\hline
            & CC~$\uparrow$	& SIM~$\uparrow$	& AUC-J~$\uparrow$	& AUC-B~$\uparrow$	 & NSS~$\uparrow$	\\
\hline \hline
CAT2000	    &\\
\hline
W-MSE	        &0.508	&0.4017	&0.8221	&0.8016	&\textbf{1.9486}\\
LC 1	        &0.5535	&\textbf{0.4261}	&0.8273	&0.8187	&1.9332\\
LC 2        &\textbf{0.5937} &0.4203 &\textbf{0.8309} &\textbf{0.8372} &1.9375\\
\hline
FiWi & \\
\hline
W-MSE	        &0.3954	&0.3872	&0.7312	&0.7114	&0.8050\\
LC 1	        &0.4118	&0.4157	&0.7621	&0.7390	&\textbf{0.8214}\\
LC 2            &\textbf{0.4236} &\textbf{0.4183} &\textbf{0.7636} &\textbf{0.7681} &0.8205\\
\hline
\end{tabular}
\end{table}

\vspace{10pt}
\textbf{SAM-VGG with linear combinations of loss functions.}

We also retrain SAM-VGG network over the training dataset, as described in Section~\ref{subsection:network}, by considering two linear combinaisons (LC 1 and LC 2). Table~\ref{table:performance_SAM_ResNet_CAT2000_FIWI} presents the results.

\begin{figure}[htbp]
    \centering
    \includegraphics[width=\linewidth]{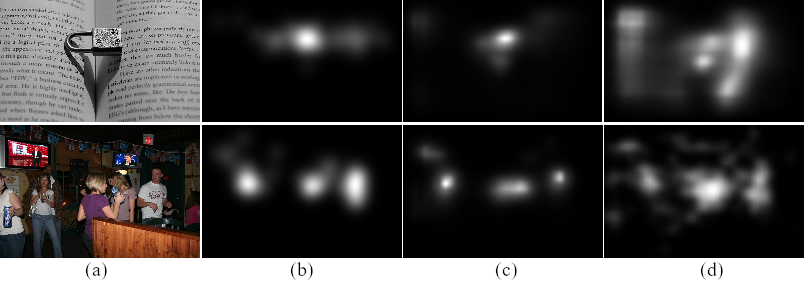}
    \caption{Example of good predictions by the combination loss while a single loss makes bad predictions (for SAM-VGG model). (a) original image; (b) Ground truth saliency map; (c) KLD + CC + NSS + DF + GM + SIG-MSE + R combination ($CC = 0.8681$ and $0.7967$); (d) MLNET-MSE ($CC=0.4320$ and $0.4491$) .}
    \label{fig:waw}
\end{figure}

The results confirm that the linear combination approach improves the prediction and hence generalizes well independently of the dataset and the network architecture. Even if the performances on the test datasets do not reach state-of-the-art techniques (see for instance~\cite{chang2018fiwi} for FiWi), due to the fact that our model was only trained on natural images in MIT, the hierarchy of the loss functions we tested remains consistent, emphasizing the benefit of the linear combination. Figure~\ref{fig:waw} illustrates a predicted saliency map generated when the SAM-VGG network is trained with a \textit{stand-alone} loss function, \ie MLNET-MSE, and with a combination of loss functions.

\begin{table}[ht!]
\small
\caption{Performance of SAM-VGG over MIT dataset with MLNET-MSE (W-MSE), KLD + CC + NSS + R (LC 1) and KLD + CC + NSS +DF +GM + SIG-MSE + R (LC 2). Best performances are in bold. (AUC-B=AUC-Borji; AUC-J=AUC-Judd)}
\label{table:performance_SAM_ResNet_CAT2000_FIWI}
\centering
\begin{tabular}{lccccc}
\hline
            & CC~$\uparrow$	& SIM~$\uparrow$	& AUC-J~$\uparrow$	& AUC-B~$\uparrow$	 & NSS~$\uparrow$	\\
\hline \hline
MIT	    &\\
\hline
W-MSE	        &0.7351	&\textbf{0.6769}	&0.8521	&0.7884	&2.0037\\
LC 1	        &0.7499	&0.6502	&0.8635	&0.7912	&\textbf{2.1694}\\
LC 2            &\textbf{0.7511} &0.6472 &\textbf{0.8712} &\textbf{0.8017} &2.0741\\
\hline
\vspace{20pt}
\end{tabular}
\end{table}

\section{Conclusion}
In this paper, we introduced a deep neural network which purpose was to evaluate the impact of loss functions on the prediction capacity when it comes to deep saliency models. We evaluated several well-known and commonly used losses, and introduced a new kind of loss function (a perceptual-based loss) that, to the best of our knowledge, has not been applied to saliency prediction. These new loss functions seem to improve, at least partially, the performances of a deep saliency model. Further work on the exact contribution of this kind of losses however still remains necessary. We showed that a simple linear combination of different losses can significantly improve over individual losses, especially when when different types of loss functions are combined (pixel-based, distribution-based, perception-based, saliency-based).

More importantly we showed that this combination strategy  generalizes well on different datasets and also with a different deep network architecture. Optimization on the coefficients of those linear combinations is also possible to obtain the best performances possible out of the combination. This approach could moreover easily be extended to other kinds of architectures, not necessarily based on convolutional neural networks.

Finally, one of the main idea that motivated our work was to highlight the importance of the choice of the loss function. We showed that a careful design of the loss function can significantly improve the performances of a model without increasing the number of trainable parameters.



\bibliography{main}

\end{document}